\pdfoutput=1

\documentclass[11pt,a4paper]{article}

\usepackage{acl}

\usepackage{times}
\usepackage{latexsym}

\usepackage[T1]{fontenc}

\usepackage[utf8]{inputenc}

\usepackage{microtype}
\usepackage{array}
\usepackage{pifont}
\usepackage{tabularx}
\usepackage{adjustbox}
\usepackage{multirow}
\usepackage{enumitem}
\usepackage{xspace}
\usepackage{tcolorbox}
\usepackage{booktabs,subcaption,amsfonts,dcolumn}
\usepackage{hyperref}
\usepackage{url}
\usepackage{amsmath,amsthm,amsfonts,amssymb,bm,stmaryrd}
\usepackage{xcolor}		
\usepackage[noorphans,vskip=0.75ex,leftmargin=2ex]{quoting}

\definecolor{darkblue}{rgb}{0, 0, 0.5}
\hypersetup{colorlinks=true, citecolor=darkblue, linkcolor=darkblue, urlcolor=darkblue}

\newcommand\ti[1]{\textit{#1}}

\newcommand\tf[1]{\textbf{#1}}
\newcommand\ttt[1]{\texttt{#1}}

\newcommand{\dtrain}{\mathcal{D}_{\text{train}}}

\newcommand{\sent}{\ttt{<}$S_1$\ttt{>}}
\newcommand{\firstsent}{\ttt{<}$S_1$\ttt{>}}
\newcommand{\secondsent}{\ttt{<}$S_2$\ttt{>}}
\newcommand{\thirdsent}{\ttt{<}$S_3$\ttt{>}}
\newcommand{\fourthsent}{\ttt{<}$S_4$\ttt{>}}

\newcommand{\template}{\mathcal{T}}

\newcommand{\vocabulary}{\mathcal{V}}
\newcommand{\labelset}{\mathcal{Y}}
\newcommand{\mapping}{\mathcal{M}}

\newcommand{\lm}{\mathcal{L}}

\newcommand{\mask}{\texttt{[MASK]}}

\renewcommand{\paragraph}[1]{\vspace{0.2cm}\noindent\textbf{#1}}

\title{Automatic Label Sequence Generation for\\Prompting Sequence-to-sequence Models}

\author{Zichun Yu$^1$ \quad Tianyu Gao$^2$ \quad Zhengyan Zhang$^{13}$ \quad Yankai Lin$^4$ \\ {\bf Zhiyuan Liu}$^{1356\dagger}$ \quad {\bf Maosong Sun}$^{1356\dagger}$ \quad {\bf Jie Zhou}$^7$ \\
$^1$Dept. of Comp. Sci. \& Tech., Institute for AI, Tsinghua University\\
$^2$Princeton University\\
$^3$Beijing National Research Center for Information Science and Technology\\
$^4$Gaoling School of Artificial Intelligence, Renmin University of China\\
$^5$International Innovation Center of Tsinghua University, Shanghai, China \\
$^6$Beijing Academy of Artificial Intelligence \\
$^7$Pattern Recognition Center, WeChat AI, Tencent Inc\\
\ttt{yuzc19@mails.tsinghua.edu.cn} \ttt{\{liuzy,sms\}@tsinghua.edu.cn}
}

\begin{document}
\maketitle


\begin{abstract}


Prompting, which casts downstream applications as language modeling tasks, has shown to be sample efficient compared to standard fine-tuning with pre-trained models. 
However, one pitfall of prompting is the need of manually-designed patterns, whose outcome can be unintuitive and requires large validation sets to tune.
To tackle the challenge, we propose \textbf{AutoSeq}, a fully automatic prompting method: 
(1) We adopt natural language prompts on sequence-to-sequence models, enabling free-form generation and larger label search space;
(2) We propose label sequences -- phrases with indefinite lengths to verbalize the labels -- which eliminate the need of manual templates and are more expressive than single label words;
(3) We use beam search to automatically generate a large amount of label sequence candidates and propose contrastive re-ranking to get the best combinations.
AutoSeq significantly outperforms other no-manual-design methods, such as soft prompt tuning, adapter tuning, and automatic search on single label words; the generated label sequences are even better than curated manual ones on a variety of tasks. 
Our method reveals the potential of sequence-to-sequence models in few-shot learning and sheds light on a path to  generic and automatic prompting. The source code of this paper can be obtained from \url{https://github.com/thunlp/Seq2Seq-Prompt}.


\end{abstract}

{\let\thefootnote\relax\footnotetext{Part of the work was done while Yankai Lin was working at Tencent.}}
{\let\thefootnote\relax\footnotetext{$^\dagger$ Corresponding authors}}


\section{Introduction}


Among ways of adapting pre-trained language models~\cite{devlin2019bert,raffel2020exploring} to downstream applications, 
prompting, which uses a natural language prompt to reformulate tasks as cloze questions, 
has shown to be especially effective~\cite{brown2020language,schick2020exploiting,schick2020size,gao2021making}. 
For example, in sentiment classification, 
prompting appends a \emph{template} ``It was \texttt{[MASK]}'' to the original input, and defines ``great'' and ``terrible'' as the \emph{label words}, whose probabilities at \texttt{[MASK]} indicate the probabilities of the positive and negative sentiment labels. 
Prompting possesses better sample efficiency and performs significantly better than standard fine-tuning in the low resource case. 







However, 
the prompting performance is highly sensitive to the prompt choice,
whose effectiveness needs abundant validation data to evaluate and is difficult to predict by intuition~\cite{gao2021making,perez2021true}. 
Even though there exist methods that explore automatic prompt search~\cite{schick2020automatically,gao2021making}, 
they still require substantial human efforts, for the algorithms start from either manual templates or label words. 

\begin{figure}[t]
\centering
\includegraphics[width=0.99\columnwidth]{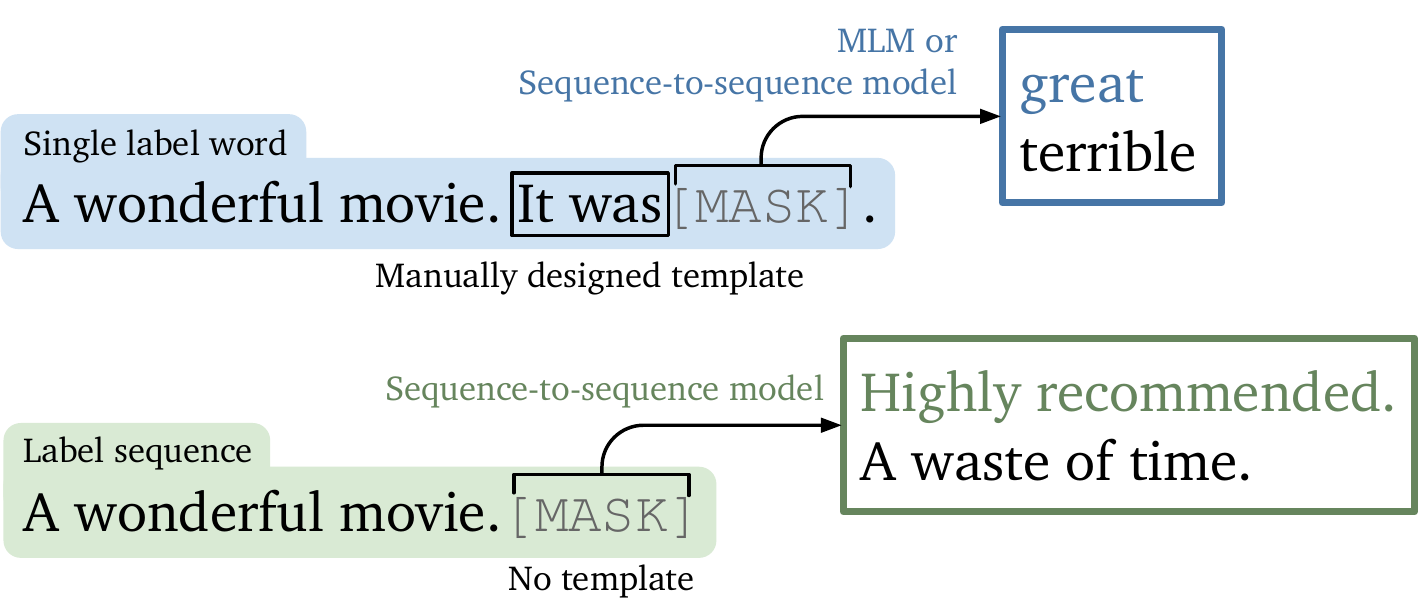}
\caption{Single label words vs label sequences. 
Label sequences are more expressive and eliminate the need of manually-designed templates.
}
\label{fig:seq_vs_one}
\vspace{-10pt}
\end{figure}




We propose AutoSeq, a prompting method that is fully automatic and requires no human input. 
AutoSeq has three innovations: 
(1) 
AutoSeq adopts \emph{sequence-to-sequence models} like T5~\cite{raffel2020exploring}.
Compared to masked language models (MLM) like BERT~\cite{devlin2019bert}, 
it allows free-form generation, enables more types of tasks, and extends the label space for prompting. 
(2) We propose \emph{label sequences}, which are indefinite-length phrases or sentences that represent each label. They are more expressive than previous single label words and eliminate the need for a manual template (Figure~\ref{fig:seq_vs_one}). 
(3) We design an automatic label sequence search pipeline, which first generates a large amount of candidates by T5, then re-ranks them by \emph{contrastive probability}. 



Our main experiment results on 
natural language understanding datasets 
show that AutoSeq performs significantly better than automatic prompt search  using single label words as well as no-prompt methods like soft prompt tuning and adapter tuning. 
AutoSeq also outperforms 
hand-crafted prompts 
on a variety of tasks. 
We hope our work enlightens automatic prompting and building a universal prompt-based fine-tuning framework.

\begin{figure*}
    \centering
    \includegraphics[width=0.98\textwidth]{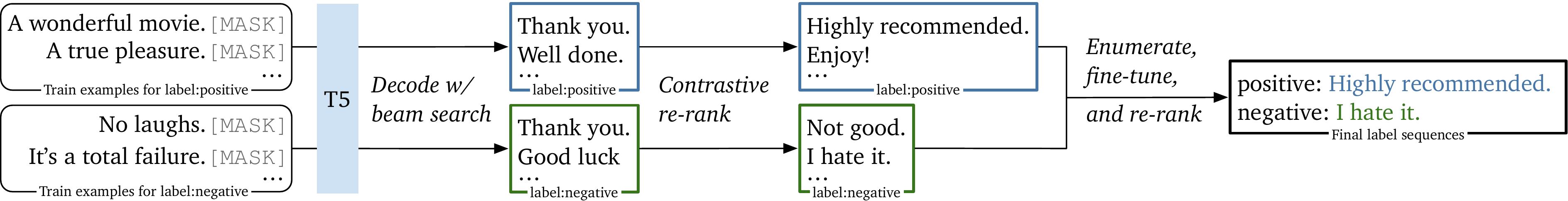}
    \caption{Illustration of AutoSeq. We first use T5 to generate label sequence candidates given each label's training instances; we then use \emph{contrastive re-ranking} to get label sequences that are more label-specific; in the end we enumerate all the combinations and re-rank by the fine-tuning performance.}
    \label{fig:auto}
    \vspace{-3pt}
\end{figure*}

\section{Related Work}


\paragraph{Prompting.}
\citet{schick2020exploiting,schick2020size,gao2021making} introduced prompting into MLM. 
Though showing remarkable few-shot performance, those models are constrained by the single \texttt{[MASK]} token and are limited to classification tasks; they also require manually-designed prompts.  
In parallel, soft prompt tuning~\cite{lester-etal-2021-power} and adapter tuning~\cite{houlsby2019parameter,zaken2021bitfit,hu2022lora} do not require manual design, but they lag behind prompting in few-shot performance~\cite{gu2022ppt}. 
Recent work~\cite{zhang2022differentiable} tries to mitigate the gap, but it still requires the help of manual prompts and thus falls out of the scope of our discussion.



\paragraph{Automatic prompt search.}
There have been plenty of attempts for automatic prompt search -- yet 
all of them require to start from either human-designed label words or templates~\cite{davison2019commonsense,jiang2020can,shin2020autoprompt,schick2020automatically,gao2021making,NEURIPS2021_e4d2b6e6,haviv-etal-2021-bertese}.
In contrast, our AutoSeq is a general-purpose, fully automatic search method that depends only on few-shot annotations.


\begin{table*}[t]
\centering
\resizebox{1.0\textwidth}{!}{%
\begin{tabular}{lcccccccc}
\toprule
& \tf{SST-2} & \tf{SST-5} & \tf{MR} & \tf{CR} & \tf{MPQA} & \tf{Subj} &  \tf{TREC} & \tf{CoLA} \\
& (acc) & (acc) & (acc) & (acc) & (acc) & (acc) & (acc) & (Matt.)\\
\midrule
Prompt tuning & 51.4 (0.0) & 24.9 (0.0) & 50.6 (0.0) & 50.7 (0.0) & 50.0 (0.0) & 59.9 (0.0) & 22.6 (0.0) & -4.0 (0.0) \\
Adapter tuning & 84.7 (3.2) & 27.7 (4.8) & 73.6 (3.9) & 86.9 (1.4) & 78.6 (3.6) & 84.7 (3.3) & 27.6 (3.6) & 4.8 (4.8) \\
\midrule
Fine-tuning & 80.9 (2.0) & 36.0 (2.2) & 70.1 (7.1) & 76.7 (6.8) & 80.9 (2.6) & 84.3 (4.4) & 68.3 (14.7) & 0.5 (6.3) \\

Prompt-based FT (Manual) & \tf{91.2} (0.7) & \tf{45.2} (1.5) & \tf{85.4} (1.5) & \tf{89.8} (1.5) & \tf{85.1} (2.9) & 89.1 (1.1) & 80.0 (2.5) & 0.7 (5.3) \\
Prompt-based FT (AutoWord) & 87.6 (2.0) & 40.4 (4.1) & 82.1 (2.7) & 87.0 (4.7) & 75.1 (4.5) & 87.4 (5.0) & 82.4 (3.7) & \underline{\tf{8.5}} (4.5) \\
Prompt-based FT (AutoSeq) & \underline{89.8} (1.1) & \underline{42.3} (3.4) & \underline{83.9} (1.3) & \underline{87.2} (2.5) & \underline{82.5} (2.7) & \underline{\tf{91.6}} (1.9) & \underline{\tf{85.2}} (4.3) & 7.6 (9.9) \\
\midrule
Fine-tuning (Full train set) & 93.3 & 56.1 & 89.3 & 86.9 & 89.0 & 96.2 & 97.0 & 30.1 \\

\midrule
    & \tf{MNLI} & \tf{MNLI-mm}  & \tf{SNLI} & \tf{QNLI} &  \tf{RTE} & \tf{MRPC} & \tf{QQP} & \tf{STS-B} \\
    & (acc) & (acc) & (acc) & (acc) & (acc) & (F1) & (F1) & (Pear.)\\
\midrule
Prompt tuning & 34.6 (0.0) & 34.2 (0.0) & 34.1 (0.0) & 54.2 (0.0) & 47.3 (0.0) & \tf{81.2} (0.0) & 53.8 (0.0) & 10.7 (0.0) \\
Adapter tuning & 33.5 (1.4) & 33.9 (1.8) & 34.7 (1.3) & 55.4 (2.5) & 50.2 (2.0) & 77.4 (2.3) & 50.7 (5.2) & 6.8 (2.8) \\
\midrule
Fine-tuning  &	36.1 (2.3) & 36.4 (2.6) & 36.0 (3.0) & 58.6 (2.5) & 51.8 (2.7) & 74.9 (5.2) & 57.0 (3.5) & 11.9 (2.8) \\

Prompt-based FT (Manual) & 41.9 (3.4) & 43.0 (3.6) & 40.8 (1.6) & 55.5 (3.1) & 53.3 (3.1) & 75.6 (7.0) & 55.4 (1.8) & 17.3 (9.5) \\
Prompt-based FT (AutoWord) & 49.0 (4.7) & 51.3 (4.6) & 56.2 (8.4) & 59.9 (4.7) & 48.7 (2.4) & \underline{73.5} (6.3) & 60.6 (4.3) & \underline{\tf{30.0}} (8.4) \\
Prompt-based FT (AutoSeq)  &	\underline{\tf{51.8}} (1.8) & \underline{\tf{53.9}} (2.0) & \underline{\tf{62.7}} (3.7) & \underline{\tf{61.3}} (4.0) & \underline{\tf{55.3}} (4.9) & 72.3 (4.9) & \underline{\tf{66.2}} (2.6) & 17.8 (13.4) \\
\midrule
Fine-tuning (Full train set) & 86.9 & 87.1 & 91.6 & 91.0 & 59.6 & 84.0 & 87.9 & 86.1 \\

\midrule
    & \tf{BoolQ} & \tf{CB}  & \tf{COPA} & \tf{MultiRC} &  \tf{ReCoRD} & \tf{WiC} & \tf{WSC} & \tf{Average} \\
    & (acc) & (F1) & (acc) & (F1) & (F1) & (acc) & (acc) & \\
\midrule
Prompt tuning & \tf{59.5} (0.0) & 36.4 (0.0) & 47.0 (0.0) & 54.4 (0.0) & 16.3 (0.0) & 50.0 (0.0) & \tf{65.4} (0.0) & 42.8 \\
Adapter tuning & 45.3 (1.5) & 55.3 (9.0) & 47.2 (3.7) & \tf{59.1} (0.0) & 23.2 (5.2) & 51.7 (1.9) & 60.2 (2.2) & 50.1 \\
\midrule
Fine-tuning  &	48.1 (6.2) & 66.4 (14.1) & 47.4 (7.2) & \tf{59.1} (0.0) & 18.1 (2.4) & 50.3 (2.8) & 60.4 (5.1) & 52.6 \\

Prompt-based FT (Manual) & 48.3 (5.5) & 75.5 (8.6) & 51.6 (1.5) & 56.0 (3.3) & \tf{56.6} (3.5) & 52.5 (3.5) & 63.5 (2.7) & 58.8 \\
Prompt-based FT (AutoWord) & 50.1 (3.9) & 66.1 (15.8) & 49.8 (3.2) & 57.9 (1.5) & \underline{\tf{56.6}} (3.5) & \underline{\tf{53.1}} (3.6) & \underline{62.1} (1.4) & 59.8 \\
Prompt-based FT (AutoSeq)  &	\underline{55.4} (8.1) & \underline{\tf{76.6}} (10.7) & \underline{\tf{52.0}} (6.8) & \underline{58.2} (0.9) & \underline{\tf{56.6}} (3.5) & 52.6 (2.9) & \underline{62.1} (1.4) & \underline{\tf{62.0}} \\
\midrule
Fine-tuning (Full train set) & 64.1 & 92.1 & 52.0 & 59.1 & 76.0 & 59.1 & 66.3 & 77.4 \\
\bottomrule
\end{tabular}
}

\caption{
Our main results using \ttt{T5-base} (16 training examples per class). 
We report mean (and standard deviation) performance over 5 different splits.
{FT:} fine-tuning;
{Manual:} human-designed prompts (Table~\ref{tab:manual_prompts});
{AutoWord:} automatically searched single label words.
The score marked as \tf{bold} means the best performance in  few-shot.
The score marked with an \underline{underline} means the best performance among automatic search methods.
}
\label{tab:main_results}
\end{table*}

\section{AutoSeq}
\label{sec:prompt}

\subsection{Prompts for sequence-to-sequence models}


We introduce the {sequence-to-sequence} version of \emph{prompt-based fine-tuning}, bringing in \ti{label sequences} that are more expressive than one token.
Using sentiment classification as an example, and given the input sentence as $x$, the model input can be formulated as ``$x$~\mask''.
We define the label sequences for the positive class as ``\emph{Highly recommended.}''
and that for the negative class as ``\emph{Not for me.}''. 
Then the probability of each class is tied with that of the T5 model generating 
``\emph{Highly recommended.}'' and ``\emph{Not for me.}'' at position $\mask$. 
As we compare the MLM single label words to our label sequences (Figure~\ref{fig:seq_vs_one}), we see that 
label sequences encode richer semantic meaning and get rid of sophisticated templates, since label sequences themselves can be standalone sentences.

In natural language inference (NLI) tasks\footnote{We have details for all tasks in Appendix~\ref{app:AutoSeq}.} with two input sentences, our model input changes to ``\text{$x_1$?~\mask, $x_2$}'' and
label sequences can be ``\emph{I mean}'' (entailment), ``\emph{For example}'' (neutral), and ``\emph{However}'' (contradiction). 

Formally, we have a task-specific template $\template$\footnote{Unlike in the MLM case, the template here is simply the way to concatenate the input and the mask.} and a task-specific mapping $\mapping \colon \labelset \rightarrow \vocabulary^+$ from the task label space $\labelset$ to the label sequence space ($\vocabulary$ is the vocabulary of the model $\lm$). 
Then, for a formulated example $\template(x)$ and its corresponding label sequences, we use the cross-entropy loss (the same way how T5 is trained)\footnote{For regression tasks like STS-B, we use the same method as \citet{gao2021making} to compute the loss instead.}
to fine-tune the model. 
In inference, we compute the score of each class $y \in \labelset$ as the auto-regressive log-probability of the corresponding label sequence:
\vspace{-0.5em}
\begin{equation}
\label{eq:lm-classification}
\resizebox{.89\hsize}{!}{$\displaystyle
    q(\mapping(y) \mid \template(x)) = \sum_{j = 1}^{|\mapping(y)|} {\log{P_{\lm}\big(t_j \mid t_{1:j-1},  \template(x)\big)}},
$}
\end{equation}
where $P_{\lm}$ denotes the output probability of the sequence-to-sequence model, $\mapping(y)=(t_1, \ldots, t_{|\mapping(y)|})$ is the corresponding label sequence tokens, and $t_{1:j-1}$ is  $t_1,...,t_{j-1}$.

\subsection{Automatic label sequence generation}
\label{sec:auto_generation}



Thanks to the introduction of label sequences, manually-designed templates are no longer needed, and the goal of automatic prompt search is simply to 
construct a label sequence mapping $\mapping$ that performs well.   
Our proposed automatic label sequence generation pipeline contains three steps (Figure~\ref{fig:auto}): 
(1) candidate generation by using T5 and beam search; 
(2) re-ranking by contrastive probability; 
(3) enumerating label sequence combinations and re-ranking by fine-tuning performance.

We first use the T5 model  and beam search to generate multiple sequence label candidates $\mathcal{S}^y \subset \vocabulary^+$ for each class $y$. 
Denote $\dtrain^y \subset \dtrain$ be the subset of all few-shot training data of class $y$, we find $s^y$ that has the top scores by this equation:
\begin{equation}
    \vspace{-1pt}
 \resizebox{.45\hsize}{!}{$\displaystyle
    \sum_{(x, y) \in \dtrain^y}q(s^y \mid \template(x)),
 $}
\vspace{-1pt}
\end{equation}
where $q(\cdot)$ is defined  as Eq. (\ref{eq:lm-classification}). Since the search space  is too large, we decompose it to an auto-regressive decoding following~\citet{gao2021making}: 
\begin{equation}
\vspace{-1pt}
    \resizebox{.73\hsize}{!}{$\displaystyle
       \sum_{j=1}^{|s^y|} \sum_{(x, y) \in \dtrain^y} \log P_{\lm}(s^y_j|s^y_{1:j-1}, \template(x)).
    $}
\vspace{-1pt}
\end{equation}
By using beam search, we can generate a large amount of label sequence candidates by just one decoding pass. 
However, 
we notice that it tends to generate similar generic label sequences across different classes, 
while we expect the label sequences to be distinguishable for each class. 
For example, in sentiment classification, both classes will get a generic candidate of ``Thank you'', which is coherent to be put at the mask but does not help with the classification (more discussion in Appendix~\ref{app:auto_gen_examples}).


\begin{table}[t]
    \begin{center}
    \centering
    \resizebox{0.87\columnwidth}{!}{%
    \begin{tabular}{lcccc}
    \toprule
            & \tf{SST-2} & \tf{SNLI} & \tf{QQP} & \tf{MultiRC} \\
    \midrule
    Manual with eng. & \tf{90.8} & \tf{64.1} & 56.1 & 57.5 \\
    AutoSeq & 89.8 & 62.7 & \tf{66.2} & \tf{58.2} \\
    \bottomrule
    \end{tabular}
    }
    \end{center}

    \caption{
        Manual prompts with engineering on large validation sets vs AutoSeq (Full results in Table~\ref{tab:manual_results_with_engineering}).
        }

    \label{tab:engineering_compare}
\end{table}

To eliminate the problem, we introduce the second step of our automatic pipeline, which re-ranks all the candidates based on the 
\emph{contrastive probability} $\tilde{q}(s^y)$ of $s^y \in \mathcal{S}^y$:
\begin{equation}
\resizebox{0.85\hsize}{!}{$\displaystyle
    \frac{\sum_{(x, y) \in \dtrain^y}q(s^y \mid \template(x))}{|\dtrain^y|} - \frac{\sum_{(x, y') \in \overline{\mathcal{D}}_{\text{train}}^y} q(s^y \mid \template(x))}{|\overline{\mathcal{D}}_{\text{train}}^y|},
$}
\end{equation}
where $\overline{\mathcal{D}}_{\text{train}}^y=\dtrain \backslash \dtrain^y$.

Then, we define the score of a label mapping as the sum of corresponding $\tilde{q}(s^y)$ for each class $y$. To shorten the time for further re-ranking, we only select the top $n$ mappings with the highest scores.
Finally, we fine-tune the model over the top $n$ label mapping candidates, and re-rank them to find the best one based on the few-shot development set, which has been proved critical in the label mapping selection \cite{gao2021making}.


\section{Experiments}

\label{sec:experiments}
\subsection{Main results}
\label{sec:mainresult}
We use a \ttt{T5-base} v1.1~\cite{shazeer2020glu}\footnote{The released original T5 models are also fine-tuned on downstream tasks while T5 v1.1 models exclude those tasks.} model and set the number of training examples per class as 16 in our experiments.
Datasets and experiments details can be found in Appendix~\ref{app:datasets} and \ref{app:exp_details}.
To make our results convincing, we compare to the following baselines in our few-shot setting:
(1) parameter-efficient tuning -- soft prompt tuning~\cite{lester-etal-2021-power} and adapter tuning~\cite{houlsby2019parameter,karimi2022perfect} -- which fixes the pre-trained model parameters and only tunes the soft prompt or adapter part;
(2) standard fine-tuning;
(3) manual prompts (Table~\ref{tab:manual_prompts}) proposed in \newcite{logan2021cutting}; 
(4) automatic label word search (AutoWord), which has the same setting as  AutoSeq  except that it is limited to only using one single token as a label word.
This can be seen as an approximation of Auto-L in \newcite{gao2021making}.
We also include the results from standard fine-tuning based on the full training set.

Table~\ref{tab:main_results} shows our main results.
First, \tf{prompt-based fine-tuning can significantly beat standard fine-tuning}, either using manual prompts or generated ones, let alone parameter-efficient tuning. 
Our  method AutoSeq 
achieves a
$9.4\%$ gain on average compared to standard fine-tuning.

Second, \tf{AutoSeq achieves a 3.2\% improvement on average compared to the manual prompts}, and performs significantly better in NLI tasks. However, for most of the sentiment classification tasks, though without engineering, the manual prompts can still outperform AutoSeq. We attribute it to the simplicity of these tasks, making the manual  design of  prompts more intuitive.

Third, \tf{using AutoSeq leads to steady gains in a majority of tasks compared to AutoWord}, indicating that label sequences, which is only enabled by using sequence-to-sequence models, are more expressive than single label words.


The results indicate that automatic prompt generation, especially with template-free format and label sequences, is a promising path for prompt-based fine-tuning in low resource scenarios.



\subsection{Analysis of prompt engineering}


Table~\ref{tab:engineering_compare} compares  manual prompts with considerable engineering efforts (Table~\ref{tab:manual_prompts_with_engineering}) to  AutoSeq.
In general, AutoSeq achieves on par performance with models using manual prompts across various types of tasks, illustrating the effectiveness of our method, especially  when trial-and-error with large validation sets is impossible.



\begin{table}[t]
    \begin{center}
    \centering
    \resizebox{0.87\columnwidth}{!}{%
    \begin{tabular}{lcccc}
    \toprule
            & \tf{SNLI} & \tf{QQP} & \tf{ReCoRD} & \tf{WSC} \\
    \midrule 
    RoBERTa-PET & 43.3 & 53.4 & 42.7 & 55.0 \\
    T5-AutoSeq & \tf{62.7} & \tf{66.2} & \tf{56.6} & \tf{62.1} \\
    \bottomrule
    \end{tabular}
    }
    \end{center}

    \caption{
        Sequence-to-sequence  vs  MLM prompting. 
        }

    \vspace{-5pt}
    \label{tab:pet}
\end{table}

\subsection{Analysis of different pre-trained models}

To highlight the advantages of using sequence-to-sequence models, we also report the PET\footnote{Without unlabeled corpora and ensemble.} 
results with \ttt{RoBERTa-base} in Table~\ref{tab:pet}.
We see that T5 performs better than RoBERTa by a large margin.
Although the comparison is not fair\footnote{Surprisingly, \ttt{T5-base} has a lower GLUE average (84.67) than \ttt{RoBERTa-base} (86.35) with full-dataset.} given T5 and RoBERTa are pre-trained with different corpora, we highlight the importance to have sequence-to-sequence models in the world of prompt-based fine-tuning.
Furthermore, for tasks like ReCoRD and WSC that require generation in prompting, T5 is perfectly fit for their output formats, while MLM models like RoBERTa require tricky workaround. 

\section{Conclusion}

In this paper, we propose AutoSeq, a prompt-based fine-tuning method with (1) sequence-to-sequence models that enable free-form generation, (2) label sequences that significantly extend the prediction space,  and (3) automatic prompt search that requires no human efforts for designing prompts.
Comprehensive experiments show that AutoSeq significantly outperforms other prompt-based or parameter-efficient tuning methods.
We hope AutoSeq further inspires research on exploring template-free prompt-based fine-tuning.

\section*{Acknowledgement}

This work is supported by the National Key Research and Development Program of China (No. 2020AAA0106500). Zichun Yu conducted the experiments. Zichun Yu, Tianyu Gao, Zhengyan Zhang, Yankai Lin, and Zhiyuan Liu wrote the paper. Maosong Sun and Jie Zhou provided valuable suggestions to the research.

\bibliography{ref}
\bibliographystyle{acl_natbib}

\clearpage
\appendix
\counterwithin{figure}{section}
\counterwithin{table}{section}

\clearpage
\section{Datasets}
\label{app:datasets}

We use datasets from GLUE~\cite{wang2019glue}, SuperGLUE~\cite{wang2019superglue}, and a number of other sentence classification datasets.

For SST-2~\cite{socher2013recursive_sst-2}, SST-5~\cite{socher2013recursive_sst-2}, MR~\cite{pang2005seeing_mr}, CR~\cite{hu2004mining_cr}, MPQA~\cite{wiebe2005annotating_mpqa}, Subj~\cite{pang2004sentimental_subj}, TREC~\cite{voorhees2000building_trec}, CoLA~\cite{warstadt2019neural_cola}, MNLI~\cite{williams2018broad_mnli}, SNLI~\cite{bowman2015large_snli}, QNLI~\cite{rajpurkar2016squad}, RTE~\cite{dagan2005pascal_rte1,bar2006second,giampiccolo2007third_rte3,bentivogli2009fifth_rte4}, MRPC~\cite{dolan2005automatically_mrpc}, QQP\footnote{\url{https://www.quora.com/q/quoradata/}} and STS-B~\cite{cer2017semeval_sts-b}, we refer to \citet{gao2021making} for their test settings.
For BoolQ~\cite{clark2019boolq}, CB~\cite{demarneffe:cb}, COPA~\cite{roemmele2011choice}, MultiRC~\cite{khashabi2018looking}, ReCoRD~\cite{zhang2018record}, WiC~\cite{pilehvar2018wic} and WSC~\cite{levesque2011winograd}, we take their original development sets as the test sets.

\section{Experimental Details}
\label{app:exp_details}

\subsection{Hyper-parameter selection}
\label{app:hyper_selection}
We take batch sizes from \{2, 4, 8\} for all few-shot experiments.
For fine-tuning, we take learning rates from \{7e-5, 1e-4, 2e-4\}.
For prompt-based fine-tuning, we take learning rates from \{2e-5, 6e-5, 9e-5\}, which are selected by pre-experiments on the SST-2 and SNLI datasets.
For each trial, we follow \citet{gao2021making} and set the training steps as 1000, validation steps as 100, then pick the best model based on the validation results.

\subsection{Automatic label sequence generation}
\label{app:AutoSeq}
For automatic label sequence generation, we use \ttt{T5-large}, limiting the maximum length of 20 tokens (AutoSeq) and one token (AutoWord).
Considering the trade-off between efficiency and effectiveness, we set beam search width to 50 and set $n$ to 20.
Given that the number of experiments is relatively large in automatic generation, we fix the batch size as 8 and the learning rate as 6e-5 when training the model over the top $n$ label mappings.

Besides our $\template$ for one-sentence classification tasks\footnote{One exception: MPQA  consists of incomplete sentences, so we adopt manual template without engineering.} and NLI tasks mentioned in Section \ref{sec:prompt}, we also design more $\template$, always a simple concatenation of input fields and the \texttt{[MASK]} token, for other complicated tasks.
For BoolQ, $\template$ is ``\text{$x_1$?~\mask, $x_2$}''.
For COPA, $\template$ is ``\text{$x_1$~$x_2$?~$x_3$?~\mask, $x_4$}''.
For MultiRC, $\template$ is ``\text{$x_2$~\mask, $x_3$~$x_1$}''.
For WiC, $\template$ is ``\text{$x_1$~$x_2$~`$x_3$'~\mask}''.
Since ReCoRD and WSC can be easily and intuitively transformed into fill-in-the-blank tasks, we follow \citet{schick2020size} and do not process the automatic label sequence generation for them.
To make the input closer to  pre-training, we refer to \citet{gao2021making} for the implementation details of prompts.

\section{Analysis of Templates}
\label{app:analysis_templates}

Table~\ref{tab:man_auto_tem} gives the results of using only manual label words with engineering and no templates (so the mask token is concatenated the same way as AutoSeq). 
This can be seen as the \ti{null prompts} from  \citet{logan2021cutting}. 
Our results further validate that null prompts perform comparably or even better to manual prompts in most cases.


\begin{table}[t]
    \begin{center}
    \centering
    \resizebox{0.87\columnwidth}{!}{%
    \begin{tabular}{lcccc}
    \toprule
            & \tf{SST-2} & \tf{SNLI} & \tf{QQP} & \tf{MultiRC} \\
    \midrule
    Manual w/o templates & 90.2 & \tf{64.1} & \tf{57.7} & 56.2 \\
    Manual with eng. & \tf{90.8} & \tf{64.1} & 56.1 & \tf{57.5} \\
    \bottomrule
    \end{tabular}
    }
    \end{center}

    \caption{
        Comparison between manual label words without templates (so the input is the same as AutoSeq), and manual prompts with
        deliberate engineering.
        }

    \vspace{-3pt}
    \label{tab:man_auto_tem}
\end{table}

\section{Manual Prompts}
\label{app:manual_prompts}

Table~\ref{tab:manual_prompts} demonstrates all the manual templates and label words adopted by us. We basically follow \citet{logan2021cutting} for these prompts. For the tasks that are not covered by \citet{logan2021cutting}, we manually write one prompt for each of them, using only our intuition.


\begin{table*}[t]
\begin{center}
\centering
\resizebox{0.99\textwidth}{!}{%
\begin{tabular}{lll}
\toprule
\tf{Task} & \tf{Template} & \tf{Label words}\\
\midrule
SST-2 &  {\sent} Overall my impression is {\mask} . & positive: good, negative: bad\\
SST-5 &  {\sent} Overall my impression is {\mask} . & v.positive: very good, positive: good, neutral: not bad, negative: bad, v.negative: very bad\\
MR    & {\sent} Overall my impression is {\mask} . & positive: good, negative: bad\\
CR    & {\sent} Overall my impression is {\mask} . & positive: good, negative: bad\\
MPQA  & {\sent} Overall my impression is {\mask} . & positive: good, negative: bad\\
Subj  & {\sent} The sentence is {\mask} . & subjective: subjective, objective: objective\\
TREC  & {\sent} The question is about {\mask} . & abbreviation: abbreviation, entity: entity, description: description\\
&& human: human, location: location, numeric: numeric\\
COLA  & {\sent} The grammar is {\mask} . & grammatical: acceptable, not\_grammatical: unacceptable\\
\midrule
MNLI  & Premise: {\secondsent} Hypothesis: {\firstsent} Label: {\mask} & entailment: yes, netural: maybe, contradiction: no\\
SNLI  & Premise: {\secondsent} Hypothesis: {\firstsent} Label: {\mask} & entailment: yes, netural: maybe, contradiction: no\\
QNLI  & Question: {\firstsent} Sentence: {\secondsent} Label: {\mask} & entailment: yes, not\_entailment: no\\
RTE   & Premise: {\firstsent} Hypothesis: {\secondsent} Label: {\mask} & entailment: yes, not\_entailment: no\\
MRPC  & {\firstsent} and {\secondsent} are the {\mask} . & equivalent: same, not\_equivalent: different\\
QQP   & {\firstsent} and {\secondsent} are the {\mask} . & equivalent: same, not\_equivalent: different\\
STS-B & {\firstsent} and {\secondsent} are the {\mask} . & $y_u$: same, $y_l$: different\\
BoolQ & Passage: {\firstsent} Question: {\secondsent} Answer: {\mask} . & True: true, False: false\\
CB    & Premise: {\firstsent} Hypothesis: {\secondsent} Label: {\mask} & entailment: yes, netural: maybe, contradiction: no\\
\midrule
COPA  & Premise: {\thirdsent} Question: {\fourthsent} Choice1: {\firstsent} & Alternative 1: Choice1, Alternative 2: Choice2\\
      & Choice2: {\secondsent} Answer: {\mask} . & \\
MultiRC  & Paragraph: {\firstsent} Question: {\secondsent} Answer: {\thirdsent} & True: true, False: false\\
         & Label: {\mask} & \\
ReCoRD  & {\firstsent} {\thirdsent} & \\
WiC   & `{\thirdsent}' in {\firstsent} and `{\thirdsent}' in {\secondsent} are the {\mask} . & True: same, False: different\\
WSC   & {\firstsent} {\thirdsent} is {\mask} . & \\
\bottomrule
\end{tabular}
}
\end{center}
\caption{Manual templates and label words following~\citet{logan2021cutting}. Note that for ReCoRD and WSC we follow \citet{schick2020size} and do not design the label words for them.
}
\label{tab:manual_prompts}
\vspace{-5pt}
\end{table*}

\section{Manual Prompts with Engineering}
\label{app:manual_prompts_with_engineering}

Table~\ref{tab:manual_prompts_with_engineering} gives all the manual templates and label words with careful engineering (\newcite{gao2021making} for GLUE and \newcite{schick2020size} for SuperGLUE) that we use in our experiments.

Table~\ref{tab:manual_results_with_engineering} compares the full results of manual prompts with engineering to our AutoSeq.
Overall, AutoSeq performs comparably or even better compared with manual prompts, particularly for tasks where developing solid manual prompts is less instinctive (e.g., TREC, QNLI, QQP and COPA).


\begin{table*}[t]
\begin{center}
\centering
\resizebox{1.98\columnwidth}{!}{%
\begin{tabular}{lll}
\toprule
\tf{Task} & \tf{Template} & \tf{Label words}\\
\midrule
SST-2 &  {\sent} It was {\mask} . & positive: great, negative: terrible\\
SST-5 &  {\sent} It was {\mask} . & v.positive: great, positive: good, neutral: okay, negative: bad, v.negative: terrible\\
MR    & {\sent} It was {\mask} . & positive: great, negative: terrible\\
CR    & {\sent} It was {\mask} . & positive: great, negative: terrible\\
MPQA  & {\sent} It was {\mask} . & positive: great, negative: terrible\\
Subj  & {\sent} This is {\mask} . & subjective: subjective, objective: objective\\
TREC  & {\mask} : {\sent} & abbreviation: Expression, entity: Entity, description: Description\\
&& human: Human, location: Location, numeric: Number\\
COLA  & {\sent} This is {\mask} . & grammatical: correct, not\_grammatical: incorrect\\
\midrule
MNLI  & {\firstsent} ? {\mask} , {\secondsent} & entailment: Yes, netural: Maybe, contradiction: No\\
SNLI  & {\firstsent} ? {\mask} , {\secondsent} & entailment: Yes, netural: Maybe, contradiction: No\\
QNLI  & {\firstsent} ? {\mask} , {\secondsent} & entailment: Yes, not\_entailment: No\\
RTE   & {\firstsent} ? {\mask} , {\secondsent} & entailment: Yes, not\_entailment: No\\
MRPC  & {\firstsent} {\mask} , {\secondsent} & equivalent: Yes, not\_equivalent: No\\
QQP   & {\firstsent} {\mask} , {\secondsent} & equivalent: Yes, not\_equivalent: No\\
STS-B & {\firstsent} {\mask} , {\secondsent} & $y_u$: Yes, $y_l$: No\\
BoolQ & {\firstsent} Question: {\secondsent} ? Answer: {\mask} . & True: Yes, False: No\\
CB  & {\firstsent} ? {\mask} , {\secondsent} & entailment: Yes, netural: Maybe, contradiction: No\\
\midrule
COPA  & {\firstsent} or {\secondsent} ? {\thirdsent} , {\fourthsent} {\mask} . & \\
MultiRC  & {\firstsent} Question: {\secondsent} Is it {\thirdsent} ? {\mask} . & True: Yes, False: No\\
ReCoRD  & {\firstsent} {\thirdsent} & \\
WiC   & {\firstsent} {\secondsent} Does {\thirdsent} have the same & True: Yes, False: No\\
      & meaning in both sentences? {\mask} & \\
WSC   & {\firstsent} The pronoun {\thirdsent} refers to {\mask} . & \\
\bottomrule
\end{tabular}
}
\end{center}
\caption{Manual templates and label words with deliberate engineering that we use in our experiments. Note that for COPA, ReCoRD and WSC, we follow \citet{schick2020size} and do not design the label words for them.
}
\label{tab:manual_prompts_with_engineering}
\vspace{-5pt}
\end{table*}


\begin{table*}[t]
\centering
\resizebox{1.0\textwidth}{!}{%
\begin{tabular}{lcccccccc}
\toprule
& \tf{SST-2} & \tf{SST-5} & \tf{MR} & \tf{CR} & \tf{MPQA} & \tf{Subj} &  \tf{TREC} & \tf{CoLA} \\
& (acc) & (acc) & (acc) & (acc) & (acc) & (acc) & (acc) & (Matt.)\\

\midrule
Manual with eng. & \tf{90.8} (0.4) & \tf{47.2} (2.4) & \tf{86.1} (0.6) & \tf{90.4} (1.0) & \tf{84.1} (2.4) & 91.4 (1.2) & 81.3 (4.8) & \tf{9.6} (11.6) \\
AutoSeq & 89.8 (1.1) & 42.3 (3.4) & 83.9 (1.3) & 87.2 (2.5) & 82.5 (2.7) & \tf{91.6} (1.9) & \tf{85.2} (4.3) & 7.6 (9.9) \\

\midrule
    & \tf{MNLI} & \tf{MNLI-mm}  & \tf{SNLI} & \tf{QNLI} &  \tf{RTE} & \tf{MRPC} & \tf{QQP} & \tf{STS-B} \\
    & (acc) & (acc) & (acc) & (acc) & (acc) & (F1) & (F1) & (Pear.)\\

\midrule
Manual with eng. & \tf{55.3} (2.3) & \tf{57.3} (2.4) & \tf{64.1} (4.1) & 59.7 (3.4) & \tf{59.1} (4.3) & 71.2 (6.6) & 56.1 (2.5) & 17.7 (12.5) \\
AutoSeq  &	51.8 (1.8) & 53.9 (2.0) & 62.7 (3.7) & \tf{61.3} (4.0) & 55.3 (4.9) & \tf{72.3} (4.9) & \tf{66.2} (2.6) & \tf{17.8} (13.4) \\

\midrule
    & \tf{BoolQ} & \tf{CB}  & \tf{COPA} & \tf{MultiRC} &  \tf{ReCoRD} & \tf{WiC} & \tf{WSC} & \tf{Average} \\
    & (acc) & (F1) & (acc) & (F1) & (F1) & (acc) & (acc) & \\

\midrule
Manual with eng. & \tf{57.5} (2.1) & \tf{79.7} (5.5) & 48.8 (2.5) & 57.5 (1.6) & \tf{56.6} (3.5) & \tf{53.4} (4.0) & \tf{62.5} (5.2) & \tf{62.5} \\
AutoSeq  &	55.4 (8.1) & 76.6 (10.7) & \tf{52.0} (6.8) & \tf{58.2} (0.9) & \tf{56.6} (3.5) & 52.6 (2.9) & 62.1 (1.4) & 62.0 \\
\bottomrule
\end{tabular}
}

\caption{
Comparison between manual prompts with engineering and our automatically searched label sequences.
}
\label{tab:manual_results_with_engineering}
\end{table*}

\section{Automatically Generated Label Sequences}
\label{app:auto_gen_examples}

We demonstrate the top 1 automatically generated label sequences before and after re-ranking with contrastive probability for all tasks in Table~\ref{tab:generated_label_sequences}. It can be observed that our contrastive probability draws a strong distinction between different classes, especially for those multi-classification tasks like SST-5 and TREC, in which our beam search tends to find the same sequence whatever the class is.

Generally speaking, the generated results after re-ranking conform with our intuition in a majority of single and two-sentence tasks. For more complicated ones, such as COPA and WiC, the generated label sequences can be counterintuitive, calling for a more elegant solution in the future.


\begin{table*}[t]
    \begin{center}
    \centering
    \resizebox{1.91\columnwidth}{!}{%
    \begin{tabular}{lll}
    \toprule
    \tf{Task} & \tf{Before re-ranking} & \tf{After re-ranking}\\
    \midrule

    \tf{SST-2} & (positive/negative) \\
     & Highly recommended./Thank you. & Highly recommended./Sigh. \\
    \midrule
    \tf{SST-5} & (very positive/positive/neutral/negative/very negative)  \\
     &  Highly recommended./Highly recommended././ \\ 
     & Highly recommended./Highly recommended. & A must see./I love this movie./Enjoy!/Sigh./Not recommended. \\
    \midrule
    \tf{MR} & (positive/negative) \\
     & Highly recommended./Highly recommended. & Highly recommended./Not for me. \\
    \midrule
    \tf{CR} & (positive/negative) \\
     & I love it./Thank you. & I love it./I hate it. \\
    \midrule
    \tf{MPQA} & (positive/negative) \\
     & ./. & ./Why? \\
    \midrule
    \tf{Subj} & (subjective/objective) \\
    & I love it./What do you think? & I love it./The rest is history. \\
    \midrule
    \tf{TREC} & (abbreviation/entity/description/human/location/numeric) \\
     & Why?/Why?/./Why?/Why?/. & Discuss!/What is it?/For?/Who is?/USA./15? \\
    \midrule
    \tf{CoLA} & (grammatical/not\_grammatical) \\
     & ./. & Enjoy!/. \\
    \midrule
    
        \tf{MNLI} & (entailment/neutral/contradiction) \\
         & Yes/Yes/No & I mean/For example/However \\
         \midrule
        \tf{SNLI} & (entailment/neutral/contradiction) \\
         & Yes/Yes/Yes & Yes/In this video/Next \\
        \midrule
        \tf{QNLI} & (entailment/not\_entailment)  \\
         & In fact/In fact & In the past/Also \\
        \midrule
        \tf{RTE} & (entailment/not\_entailment) \\
         & Yes/Yes & Yes/However \\
        \midrule
        \tf{MRPC} & (equivalent/not\_equivalent) \\
         & Yes/Yes & Yes/Meanwhile \\
        \midrule
        \tf{QQP} & (equivalent/not\_equivalent)\\
         & Also/Also & So/Also \\
        \midrule
        \tf{STS-B} & ($y_u$/$y_l$)\\
         & Yes/Yes & Yes/Also \\
        \midrule
        \tf{BoolQ} & (True/False)\\
         & Yes/Yes & If so/No \\
        \midrule
        \tf{CB} & (entailment/neutral/contradiction) \\
         & Yes/Yes/I mean & Indeed/A: Yes/A: No \\
        \midrule
        \tf{COPA} & (Alternative 1/Alternative 2)\\
         & No/No & No/Yes \\
        \midrule
        \tf{MultiRC} & (True/False)\\
         & Yes/Yes & The answer is/Also \\
        \midrule
        \tf{WiC} & (True/False)\\
         & is used./is used. & is used./is an adjective. \\
        \bottomrule
    
    \end{tabular}
    }
    \end{center}
    \caption{Top 1 automatically generated label sequences before and after re-ranking with contrastive probability for all tasks based on one few-shot split.}
    \label{tab:generated_label_sequences}
\end{table*}

\end{document}